\pdfoutput=1
\documentclass[11pt,a4paper]{article}
\usepackage{ranlp2023}

\aclfinalcopy % Uncomment this line for the final submission
\def\aclpaperid{116} %  Enter the acl Paper ID here

\usepackage{times}
\usepackage{latexsym}
\usepackage{booktabs}
\usepackage{enumitem}
\usepackage{cleveref}
\usepackage{multirow}

\usepackage[T1]{fontenc}
\usepackage[utf8]{inputenc}

\usepackage{microtype}

\usepackage{inconsolata}

\usepackage{graphicx}

\usepackage[normalem]{ulem} % sout, uline
\usepackage{xcolor}
\title{LLM Compression: How Far Can We Go in Balancing Size and Performance?}

\author{
\textbf{Sahil Sk}$^{*}$, \textbf{Debasish Dhal}$^{*}$, \textbf{Sonal Khosla}$^{*}$, \textbf{Sk Shahid}$^{*}$,\\
\textbf{Sambit Shekhar}$^{*}$, \textbf{Akash Dhaka}$^{\dagger}$, \textbf{Shantipriya Parida}$^{\dagger}$,\\
\textbf{Dilip K. Prasad}$^{\ddagger}$, \textbf{Ond\v{r}ej Bojar}$^{\S}$\\[2mm]
$^{*}$Odia Generative AI, India;
$^{\dagger}$AMD Silo AI, Finland\\
$^{\ddagger}$The Arctic University of Norway, Norway\\
$^{\S}$Charles University, MFF, \'{U}FAL, Czech Republic\\[2mm]
\texttt{Correspondence: akash.dhaka@amd.com}
}

\begin{document}
\maketitle
\begin{abstract}
Quantization is an essential and popular technique for improving the accessibility of large language models (LLMs) by reducing memory usage and computational costs while maintaining performance. In this study, we apply 4-bit Group Scaling Quantization (\textit{GSQ}) and Generative Pretrained Transformer Quantization (\textit{GPTQ}) to \textit{LLaMA 1B}, \textit{Qwen 0.5B}, and \textit{PHI 1.5B}, evaluating their impact across multiple NLP tasks. We benchmark these models on \textit{MS MARCO }(Information Retrieval), \textit{BoolQ} (Boolean Question Answering), and \textit{GSM8K} (Mathematical Reasoning) datasets, assessing both accuracy and efficiency across various tasks. The study measures the trade-offs between model compression and task performance, analyzing key evaluation metrics 
namely: accuracy, inference latency, and throughput (total output tokens generated per second), providing insights into the suitability of low-bit quantization for real-world deployment. %\repl{ and highlight the tradeoffs between memory, computing and latency in such settings, helping a user make suitable decisions}{}\XXX{I still think this is too similar to the previous list in the sentence and should be removed}.
Using the results, a user can then make a suitable decision based on the specifications that need to be met. We discuss the pros and cons of GSQ and GPTQ techniques on models of different sizes, which also serve as a benchmark for future experiments.

\end{abstract}

\section{Introduction}

The increasing demand for high-performing LLMs has driven the development of transformer architectures with billions of parameters, capable of achieving state-of-the-art results and unlocking new capabilities in various language understanding tasks such as reasoning, proof-checking, and automated software development. 
%Recent research has unravelled the so-called empirical scaling laws for language model pretraining which have so far come to the consensus that proportionally increasing the model capacity along with computation budget and training dataset size consistently decreases perplexity, improves performance on downstream tasks like commonsense reasoning, software development, etc. and thereby achieves the highly sought-after emergent capabilities \cite{srivastava2022beyond}. 
However, the size and complexity of these models often pose significant challenges, including high computational costs in terms of floating point operations per second, memory requirements, and energy consumption or limited throughput. These limitations hinder the deployment of such models on resource-constrained devices such as mobile phones, Internet of Things (IoT) devices, and edge computing platforms.

Quantization is one of the techniques that has gained prominence lately; it reduces the precision of model weights and activations, for efficient deployment on resource-constrained hardware. It reduces the number of bits used to represent each parameter (e.g., from 32-bit floating-point, FP32, to 8-bit integer, INT8, or lower), thereby enabling lower memory usage and faster inference. 

 %In many scenarios, it has been seen that quantization outperforms \cite{kuzmin2024pruningvsquantizationbetter} pruning and knowledge distillation in terms of reducing the model size while maintaining performance in terms of quality.
 Quantization can also support deployment on low resource/power devices like FPGA, Neural Processing Units (NPU) and System on a chip (SOCs) and can be combined easily with other compression techniques like knowledge distillation and pruning and multiply compression effects.
 %For such systems, compute is not the only constraint, but memory bandwidth and throughput also serve as important desiderata.
 %Since it works orthogonally with respect to the other techniques, it can be combined with them to reduce model size and memory requirements. \cite{li2023model}. 
%Model compression has also led to efficient inference in large language models \cite{wang2024model}. 

%While model capacity is often quantified by the number of model parameters the better way to do this is to use FLOPS \textit{compute per example} which is the amount of floating point operations performed for a given fixed length input.    

%LLM performance, in terms of output quality, %\XXX{unclear, if you refer to performance in terms of quality or in terms of processing speed: Shantipriya: output quality } 
%has also been shown to be directly proportional to the extent of model compression \cite{yin2024entropy}. Quantization techniques aim to reduce model size while preserving performance. 
% In this paper, we explore the effectiveness of \repl{GSQ and GPTQ compression techniques}{two compression techniques, GSQ and GPTQ,} in computing the resource requirements of a model, and their impact on various performance metrics. 
In this paper, we explore the effectiveness of two compression techniques, GSQ and GPTQ, in computing the resource requirements of a model as well as their impact on various performance metrics. 

Based on our experiments, our key findings are as follows:
\begin{itemize}

\item We observe that whether to use quantization and the choice of technique will ultimately depend on the user's requirements in terms of tasks and the model they decide to use. 
%On reasoning tasks, quantization does result in significant drop in accuracy or perplexity, on the other hand for some models and tasks, quantization produces no significant changes in accuracy. 

\item 4-bit quantization schemes used in this work had little to no impact on latency and throughput, supporting their practical deployment on production. In some cases, there was a noticeable overhead due to their implementation.

\end{itemize}

\section{Related Work}

%There are broadly two types of Quantization, \textit{Post training Quantization (PTQ)} and \textit{Quantization Aware Training (QAT)} depending upon when and how quantization is applied during the model development process. '

%The quantization technique we work with in this paper, Post-training quantization (PTQ) is applied to the model in floating-point precision (typically FP32). The model is then converted into a lower-bit representation (e.g., INT8, 4-bit) without retraining. Instead of training a model in full precision and quantizing it afterward, QAT simulates quantization during training, allowing the model to adjust to lower-bit inference\cite{hasan2024optimizinglargelanguagemodels}. %Training a DNN in low-precision usually results in worse
%performance compared to training in full precision.
%Although the QAT method has higher accuracy retention than PTQ, as the model learns to adjust to lower bit precision, it is a computationally expensive method, as it requires full retraining \cite{hasan2024optimizinglargelanguagemodels} also facing convergence issues when the loss landscape of the network is highly multimodal.  

Some of the Post-training quantization (PTQ) techniques are: static quantization (converts both weights and activations to a lower-bit format (e.g., INT8; \cite{montestruque2007static}), dynamic quantization (only weights are quantized (e.g., FP32 $\rightarrow$ INT8), while activations remain in FP32; \cite{montestruque2007static}), weight-only quantization (only model weights are quantized; \cite{kim2023finequant}), GSQ which splits model weights into small groups and applies different scaling factors per group \cite{zeng2025gqsagroupquantizationsparsity}, GPTQ which quantizes a LLM one layer after another \cite{sharify2024posttrainingquantizationlarge}, KL Divergence Based Quantization \cite{xie2016kl} and Smooth Quantization \cite{xiao2024smoothquantaccurateefficientposttraining}.

\section{Methodology}

\subsection{Quantization Techniques}
% In this section, we discuss the mechanism behind the two techniques used \repl{to compress the three transformer-based language models described above}{in our experiments}\XXX{I don't think the models were introduced above}. 
In this section, we discuss the mechanism behind the two techniques used in our experiments.

%\subsubsection{GPTQ (4-bit)}
%GPTQ is a Post training Quantization technique.  A detailed working of the GPTQ is given in Figure 1. 

%\begin{figure*}[h]
%    \centering
%    \includegraphics[width=1\textwidth]{latex/figures/GPTQ.png}
%    \caption{GPTQ 4-bit Quantization Process}
%    \label{fig:gptq}
% s\end{figure*}

\subsubsection{Generative Pretrained Transformer  Quantization (GPTQ)}
GPTQ is a one-shot quantization technique that reduces the model size by converting weights to a lower bit representation (such as 8-bits or 4-bits) from the original 32 bit or 64 bit precision \cite{frantar2022gptq}. Since this could lead to a loss of model accuracy, GPTQ minimizes quantization errors using a dynamic error correction technique that adjusts subsequent weights to compensate for previous errors during inference.

This also allows for faster computation during inference \cite{10628367}, as lower-precision arithmetic operations (e.g., 8-bit multiplications) are more computationally efficient than high-precision operations.
% However, this might itself be offset by the extra layer of computation required to change the precision.
%A detailed working of the GPTQ technique is shown in Figure...

\subsubsection{Group Scaling Quantization (GSQ)}
\begin{figure*}[!htb]
    \centering
    \includegraphics[width=\textwidth]{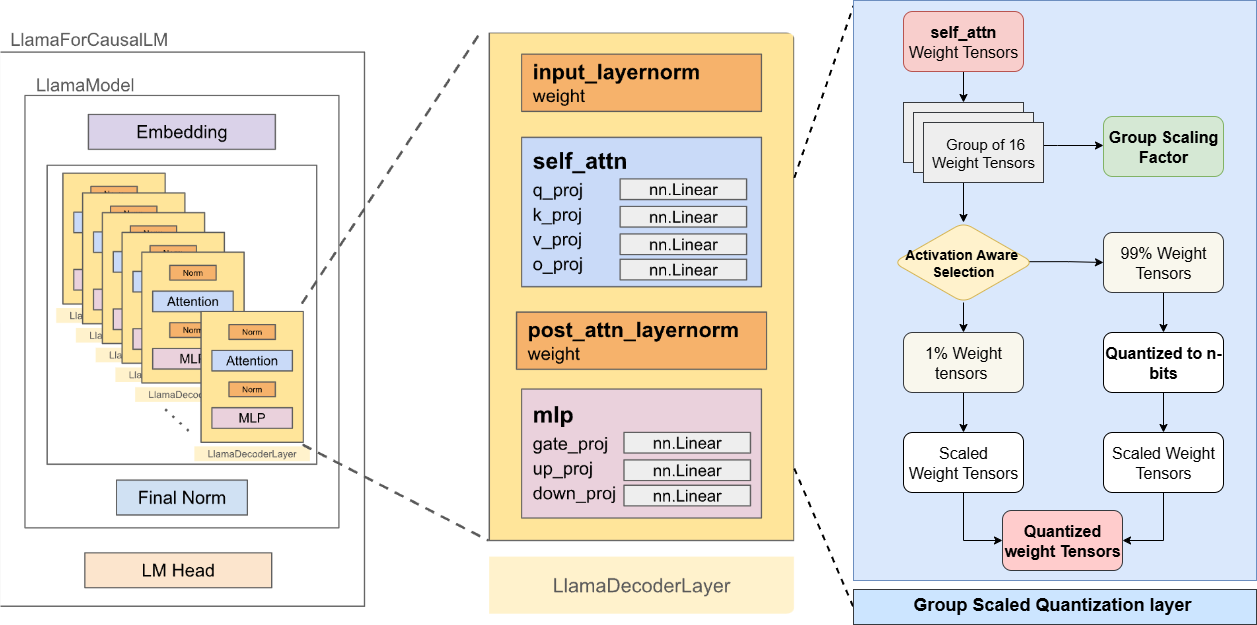}
    \caption{Detailed view of the architecture behind GSQ Quantization Process.}
    \label{fig:gptq}
\end{figure*}

GSQ, is based on Activation Weight Quantization (AWQ)~\citep{lin2024awq} introduces an innovative technique that prioritizes activation-aware scaling,
% scaling, ensuring high model fidelity while significantly reducing computational overhead. 
% % This technique helps to reduce the computational overhead by allowing for efficient quantization without needing a unique scaling factor for every individual weight. 
% To balance efficiency and accuracy, 
GSQ divides the weight matrix into groups and assigns a shared scaling factor. This ensures that all quantized values fit within the INT4 range, minimizing precision loss. Instead of selecting weights based on magnitude, GSQ identifies important weights by quantifying their impact on activation. GSQ has been shown to preserve more fine-grained information and potentially yield higher post-quantization accuracy. The working of the technique is shown in Figure \ref{fig:gptq}. 

% \begin{figure*}[!htb]
%     \centering
%     \includegraphics[width=\textwidth]{latex/figures/GSQ-drawio.png}
%     \caption{Detailed view of the architecture behind GSQ Quantization Process}
%     \label{fig:gptq}
% \end{figure*}

\subsection{Experiment Setup and Metrics}
\label{sec:exp-setup-metrics}

In this section, we describe our experimental configuration and the benchmark datasets used. We evaluated three language models to assess the impact of quantization and group size on various downstream tasks:

% \begin{itemize}[noitemsep]
%     \item LLaMA 3.2-1B
%     \begin{itemize}
%         \item Fine-tuned on \textit{GSM8K} and \textit{SQuAD}.
%         \item Validated on \textit{BOOLQ}, \textit{MS MARCO}, and \textit{GSM8K} with 5,000 samples.
%         \item Pre- and post-quantization results are reported in Table~1.
%     \end{itemize}
%     \item Qwen 0.5B
%     \begin{itemize}
%         \item Fine-tuned on \textit{SQuAD}.
%         \item Validated on \textit{BOOLQ} and \textit{MS MARCO} with 5,000 samples.
%         \item Pre- and post-quantization results are reported in Table~2.
%     \end{itemize}
%     \item Phi 1.5B
%     \begin{itemize}
%         \item Fine-tuned on \textit{GSM8K} and \textit{SQuAD}.
%         \item Validated on \textit{BOOLQ}, \textit{MS MARCO}, and \textit{GSM8K} with 5,000 samples.
%         \item Pre- and post-quantization results are reported in \textit{Table~3}.
%     \end{itemize}
% \end{itemize}

\begin{itemize}[noitemsep, topsep=0pt, leftmargin=*]
    \item LLaMA 3.2-1B~\citep{llama2}: Fine-tuned on GSM8K ~\citep{cobbe2021}; validated on BOOLQ~\citep{clark-etal-2019-boolq}, MS MARCO~\citep{nguyen2016}, GSM8K (5,000 samples). 
    \item Qwen 0.5B~\citep{qwen2}: Fine-tuned on; validated on BOOLQ, MS MARCO (5,000 samples). 
    \item Phi 1.5B~\citep{phi}: Fine-tuned on GSM8K; validated on BOOLQ, MS MARCO, GSM8K test split(5,000 samples).
\end{itemize}

%For experiments using GSQ, the parameters of each model were quantized into 16 and 8 groups per layer using a 4-bit representation, but the results of only the 16-group size have been reported in this paper. 
Results are reported for both pre-quantization and post-quantization stages. 

\begin{enumerate}[noitemsep]
    \item Pre-quantization (Unquantized) Evaluation
    %We first evaluate the original on the chosen validation datasets.
    \item Post-quantization (Quantized) Evaluation: We then apply quantization, maintaining the same validation setup to measure changes in accuracy and other metrics.
\end{enumerate}

To thoroughly compare pre-quantization and post-quantization model performance, we used a combination of accuracy based metrics like perplexity, and accuracy and efficiency based metrics like inference latency, throughput, and memory usage. 
Accuracy evaluates the proportion of correct predictions in binary classification tasks as in dataset like BoolQ, while perplexity measures the log-likelihood score at each generation step.

%Perplexity is a standard metric in language modeling that indicates how well a model predicts the next token, getting a log-likelihood score at each generating step and then summing it up before normalizing for sentence length. Lower perplexity and correspondingly higher log likelihood implies better predictive performance and language understanding \cite{Jelinek1977PerplexityaMO}. It is good primarily for tasks focussing on generative or completion objectives (e.g., GSM8K). Accuracy evaluates the proportion of correct predictions in classification or span-prediction tasks \cite{powers2020evaluationprecisionrecallfmeasure}. We report results on BOOLQ, MS MARCO, GSM8K, and SQuAD validation sets with the corresponding metrics to capture how well the model addresses classification on QA tasks.

Inference Latency is the average time needed to generate a response or process a batch of inputs \cite{hasan2024optimizinglargelanguagemodels}. By comparing pre and post-quantization latency, we determine the impact of quantization on real time or near-real-time deployment scenarios, taking into account any effect due to the overhead involved. Throughput measures how many samples (or tokens) a model can process per second~\cite{jouppi2017indatacenterperformanceanalysistensor}. We compare the rate at which models can handle input data in both unquantized and quantized forms, illustrating the trade-offs between speed and accuracy. Memory usage tracks the RAM and GPU memory footprint during inference \cite{frantar2022gptq}. Lower memory usage can enable deployment on resource-constrained devices. We quantify the difference in memory usage before and after quantization to show whether quantization yields tangible resource savings. 

All experiments were conducted on one NVIDIA A100 SXM4 40GB GPU using PyTorch and the Transformers libraries.
%Sample outputs can be found in Table \ref{tab:sample-table} (Appendix).

\section{Results and Analysis }
The results are summarized and analyzed in Table \ref{tab:quantization-summary}. For the LLaMA 1B model, GSQ actually improved accuracy on MS MARCO (81.12\%~$\rightarrow$~84.04\%) with minimal impact on latency, memory, or throughput. GPTQ maintained high accuracy, with a small reduction in memory usage after quantization. On BoolQ, GSQ significantly increased throughput (69.57~$\rightarrow$~364.91), while accuracy remained mostly unchanged for both. On GSM8K, GSQ showed a slight accuracy drop (1.21\%~$\rightarrow$~1.14\%) with a small increase in memory use.

Overall, GPTQ outperforms GSQ in accuracy for MS MARCO and BoolQ, while GSQ maintains better stability in latency and throughput but underperforms on GSM8K.

For the \textit{QWEN 0.5B} model, GSQ's accuracy declined (19.54\%~$\rightarrow$~14.90\%), while GPTQ improved (6.84\%~$\rightarrow$~10.66\%). On BoolQ, accuracy changes were minor, indicating low sensitivity to quantization. Latency increased for both methods (0.1105~s~$\rightarrow$~0.2459~s). Throughput dropped notably across GSM8K, MS MARCO, and BoolQ.

In summary, quantization affects throughput and latency more than accuracy, with GPTQ offering better accuracy overall, and GSQ delivering more stable runtime performance.

\begin{table*}[!htb]
    \centering
    \small

    \begin{tabular}{|p{2.0cm}|p{1.5cm}|p{2.2cm}|p{2.2cm}|p{2.2cm}|p{2.5cm}|}
    \hline
    \textbf{Model} & \textbf{Dataset} & \textbf{Baseline Score (\%)} & \textbf{Best Quantized Score (\%)} & \textbf{Memory Reduction} & \textbf{Key Observations} \\ \hline
    
    \multirow{3}{*}{\textbf{LLama 1B}} 
    & GSM8K & 1.21 & 1.14 (GSQ) & -391 MB$^\dagger$ & GPTQ performance score is very low 
    \\ \cline{2-6}
    & MS MARCO & 81.12 & \textbf{99.86 (GPTQ)} & +2691 MB$^\dagger$ & GSQ improves baseline \\ \cline{2-6}
    & BoolQ & 40.15 & \textbf{62.17 (GPTQ)} & +2240 MB$^\dagger$ & Significant improvement \\ \hline
    
    \multirow{3}{*}{\textbf{QWEN 0.5B}} 
    & GSM8K & -- & 0.00 (GPTQ) & No change & Both methods fail/missing \\ \cline{2-6}
    & MS MARCO & 19.54 & \textbf{14.90 (GSQ)} & No change & All scores degraded \\ \cline{2-6}
    & BoolQ & 56.21 & \textbf{55.81 (GSQ)} & No change & Minimal degradation \\ \hline
    
    \multirow{3}{*}{\textbf{Phi 1.5B}} 
    & GSM8K & 2.58 & 2.50 (GSQ) & +527 MB$^\dagger$ & GPTQ performance score is very low \\ \cline{2-6}
    & MS MARCO & 99.80 & \textbf{99.82 (GSQ)} & -2742 MB & Slight improvement \\ \cline{2-6}
    & BoolQ & -- & 40.18 (GPTQ) & -437 MB & Baseline missing \\ \hline
    \end{tabular}
    
    \vspace{0.3cm}
    
    \begin{tabular}{|p{3.0cm}|p{10.0cm}|}
    \hline
    \textbf{Critical Issues} & \textbf{Description} \\ \hline
    \textbf{GPTQ Math Failure} & GPTQ achieves very low accuracy on GSM8K across ALL models  \\ \hline
    \textbf{Memory Paradox} & Quantized models often use MORE memory than baseline (marked with $^\dagger$) \\ \hline
    \textbf{Missing Data} & Extensive GSQ data missing, particularly for BoolQ datasets \\ \hline
    % \textbf{Perfect Score Anomaly} & All models achieve exactly 100\% on SQUAD (marked with $^*$) - likely measurement error \\ \hline
    \textbf{Inconsistent Benefits} & Quantization benefits vary dramatically by model-dataset combination \\ \hline
    \end{tabular}
    
    % \caption{Executive Summary: Quantization Performance Across Models. \repl{$^*$ Perfect scores indicate likely measurement errors.}{} $^\dagger$ Negative memory reduction (quantized uses more memory) suggests experimental issues or inefficient implementation. Best performing quantization method shown in bold.}
    \caption{Executive Summary: Quantization Performance Across Models. $^\dagger$ Negative memory reduction (quantized uses more memory) suggests experimental issues or inefficient implementation. Best performing quantization method shown in bold. The memory footprint for each dataset is different due to different input token lengths.}
    \label{tab:quantization-summary}
    
\end{table*}

% \begin{figure*}[!htb]
%             %\includegraphics[width=.3\textwidth]{latex/figures/qwen_0.5b_performance.png}\hfill
%             \includegraphics[width=.49\textwidth]{latex/figures/phi1_1.45b_performance.png}\hfill
%             \includegraphics[width=.49\textwidth]{latex/figures/llama3_2.1b_performance.png}
%             \caption{Perplexity scores Pre and Post Quantization using GSQ (Data from Table \ref{tab:perplexity-scores-combined})}
%             \label{fig:spider-gsq}
% \end{figure*}

The \textit{Phi} model is a small-scale language model capable of general NLP tasks and QA tasks. Although they are trained on GSM8K, the accuracy is low and also there is hardly any difference after quantization. The model struggles with GSM8K regardless of quantization. Quantization had minimal impact on inference latency, and memory remained almost unchanged across models. 

However, it is worth noting is that there is a significant drop in throughput across all datasets in the GPTQ method, although the GSQ method still does not have a major drop in throughput. 
While \textit{4-bit quantization} at a \textit{16-group size} can lead to higher accuracy retention, the above experiments revealed the following trade-offs when compared to larger-group quantization:

\begin{itemize}[noitemsep]
    \item Improved Accuracy: The structured quantization approach preserves crucial model parameters more effectively, enhancing accuracy on validation tasks.
    \item Reduced model size: The quantization methods are able to achieve  up to a 13-fold reduction in model size with minimal drop in performance across all benchmarks.
    \item Increased Latency: Processing smaller groups introduces additional overhead, resulting in slightly higher per-inference execution time.

\end{itemize}

These observations underscore the importance of balancing group size, bit precision, and task requirements. For deployments that prioritize accuracy, a 16-group size may be ideal despite the higher cost in latency, throughput, and memory usage.

Perplexity has slightly increased after quantization across all three models for Wikitext, MS MARCO, BoolQ, and GSM8K. The increase in perplexity is more pronounced in GPTQ than in GSQ. In edge cases, GPTQ gives us an almost two-fold increase in perplexity, while for GSQ, no such event was observed.

In summary, we examined the effects of \textit{GSQ-based 4-bit quantization} and \textit{GPTQ-based 4-bit quantization} on three different language models: \textit{LLaMA 3.2--1B}, \textit{Qwen 0.5B}, and \textit{Phi 1.5B} across various tasks. The \textit{16-group size} configuration generally preserved higher accuracy at the expense of increased latency, lower throughput, and elevated memory usage. The metrics such as \textit{perplexity}, \textit{accuracy}, \textit{inference latency}, \textit{throughput}, and \textit{memory usage} provided the trade-offs faced when compressing large language models. 

The experiments for GSQ were also tried with a group size of 64 and a maximum sequence length of 512.  
%In supplementary material, we present results for inference latency, throughput, and memory usage. 
The throughput has increased across all models. Worth noting is that there is a drop in latency post quantization in all models, though the drop is very small for LLama.

\section{Limitations}
We highlight key limitations of GSQ (and partially GPTQ) observed during our experiments:

\begin{itemize}[noitemsep]

    \item \textit{Group Size Constraint:}  
    GSQ requires the last tensor dimension to be divisible by the group size (e.g., a (256, 100) tensor fails with group size 32 due to misalignment).

    \item \textit{Lack of Layer-wise Flexibility:}  
    A fixed group size across layers restricts GSQ's applicability to models with varying layer dimensions.

    \item \textit{Sensitivity to Fine-Tuning:}  
    Fine-tuned models often introduce sparsity or minor structural changes. Large group sizes tend to fail; smaller ones ($\leq16$) work better.

    \item \textit{No Fallback Handling:}  
    GSQ lacks mechanisms to detect or adapt to incompatible shapes, leading to runtime failures.

\end{itemize}

\section{Conclusion and Future work}
Our results show that LLaMA 1B benefits from quantization, even outperforming the base model on MS MARCO, while smaller models like Qwen 0.5B suffer significant accuracy loss. BoolQ remains largely unaffected by quantization, whereas GSM8K, a math-focused dataset, demonstrates sensitivity due to precision loss. Efficiency metrics reveal minimal impact on latency and throughput, suggesting that 4-bit quantization is a viable compression technique for real-world deployment. Future work includes a layer-wise analysis of the effects of 4-bit quantization.

\section*{Acknowledgement} This work was supported by the Research Council of Norway Project (nanoAI, Project ID: 325741).

\bibliographystyle{plain}
\bibliography{references}

%\newpage

\end{document}